\documentclass{article}

\usepackage{arxiv}

\usepackage[utf8]{inputenc} 
\usepackage[T1]{fontenc}    
\usepackage{hyperref}       
\usepackage{url}            
\usepackage{booktabs}       
\usepackage{amsfonts}       
\usepackage{nicefrac}       
\usepackage{microtype}      
\usepackage{graphicx}
\usepackage{natbib}
\usepackage{doi}
\usepackage{multirow}
\usepackage{xcolor}
\usepackage{color}
\usepackage{pdfrender}

\newcommand{\xmark}{x}
\newcommand{\redx}{\textcolor{red}{\xmark{}}}
\definecolor{nicergreen}{RGB}{57,151,92}
\newcommand*{\boldcheckmark}{%
  \textpdfrender{
    TextRenderingMode=FillStroke,
    LineWidth=.75pt, 
  }{\checkmark}%
}
\newcommand{\greencheck}{\textcolor{nicergreen}{\boldcheckmark{}}}

\title{DeepSpeed Ulysses: System Optimizations for Enabling Training of Extreme Long Sequence Transformer Models}

\author{Sam Ade Jacobs, Masahiro Tanaka, Chengming Zhang, Minjia Zhang \And
Shuaiwen Leon Song, Samyam Rajbhandari, Yuxiong He \\
Microsoft Inc \\
\texttt{\{samjacobs,mtanaka,minjiaz,v-chengmingz,leonsong,samyamr,yxhe\}@microsoft.com}}

\date{}

\renewcommand{\headeright}{}
\renewcommand{\undertitle}{}
\renewcommand{\shorttitle}{\textit{DeepSpeed-Ulysses}}

\hypersetup{
pdftitle={DeepSpeed Ulysses: System Optimizations for Enabling Training of Extreme Long Sequence Transformer Models},
pdfsubject={q-bio.NC, q-bio.QM},
pdfauthor={Sam Ade Jacobs, Masahiro Tanaka, Chengming Zhang, Minjia Zhang, Leon Song, Samyam Rajbhandari, Yuxiong He},
}

\begin{document}

\maketitle

\begin{abstract}
 Computation in a typical Transformer-based large language model (LLM) can be characterized by batch size, hidden dimension, number of layers, and sequence length. Until now, system works for accelerating LLM training have focused on the first three dimensions: data parallelism for batch size, tensor parallelism for hidden size and pipeline parallelism for model depth or layers. These widely studied forms of parallelism are not targeted or optimized for long sequence Transformer models. Given practical application needs for long sequence LLM, renewed attentions are being drawn to sequence parallelism. However, existing works in sequence parallelism are constrained by memory-communication inefficiency, limiting their scalability to long sequence large models. In this work, we introduce DeepSpeed-Ulysses, a novel, portable and effective methodology for enabling highly efficient and scalable LLM training with extremely long sequence length. DeepSpeed-Ulysses at its core partitions input data along the sequence dimension and employs an efficient all-to-all collective communication for attention computation. Theoretical communication analysis shows that whereas other methods incur communication overhead as sequence length increases, DeepSpeed-Ulysses maintains constant communication volume when sequence length and compute devices are increased proportionally. Furthermore, experimental evaluations show that DeepSpeed-Ulysses trains 2.5x faster with 4x longer sequence length than the existing method SOTA baseline.  
\end{abstract}


\section{Introduction}
Training large models with long sequences is becoming very important across the board from generative AI to models for scientific discovery. On generative AI side, conversational AI, knowledge-rich long document summarization and video generation require reasoning over long contexts in spatial and temporal domains. For example, multimodal foundation models such as ones that process speech, images and waveforms concurrently require long context reasoning over high dimensional inputs with long sequences. Similarly, chapter and book level summarization (estimated at tens and hundreds of thousands of words) are of great importance in conversational AI and abstractive summarization tasks ~\citep{beltagy2020longformer,kryściński2022booksum,mpt7B} and have shown to benefit from long sequence training ~\citep{xiong2023effective,peng2023yarn,touvron2023llama}. The debut of ChatGPT (and subsequent similar open source and "product" LLM brands) has pushed chat application to the forefront of modern AI, making chat applications to be more relevant than ever before. Processing long sequence is crucial for supporting longer histories in chat applications ~\citep{touvron2023llama}.

Long sequence length is equally critical for AI for science opening doors for better understanding of structure biology, health care, climate and weather forecasting ~\citep{nguyen2023climax} and large molecular simulation ~\citep{zvyagin2022genslms}. For instance, by adapting large language models with gene sequences, we can create language models that can learn the evolutionary patterns of genomes using simple alphabets and extremely long sequences (the human genome has 6.4 billion letters) ~\citep{zvyagin2022genslms}. In health care, diagnostic predictive model conditioned on entire patient care record requires context of long sequences ~\citep{DBLP:journals/corr/abs-2201-11838,9364676}.

Despite the emerging importance of long sequence length for both generative AI and AI for science, existing large model training systems and the underlying parallelism technologies (data, tensor, pipeline, sequence parallelism) are limited in their ability to support the efficient long sequence training. Two challenges with existing parallelism approach come to the fore. First, existing parallelism approach such as data, tensor and pipeline parallelism cannot address the scaling along sequence dimension. Second, existing sequence parallelism approaches are not effective because of memory-communication inefficiencies. Furthermore, existing approaches have limited usability requiring intrusive and error prone code refactoring.

In this paper, we introduce DeepSpeed-Ulysses (or Ulysses, a very long novel), a simple, portable, and effective methodology for enabling highly efficient and scalable LLM training with extremely long sequence lengths. 
DeepSpeed-Ulysses partitions individual samples along the sequence dimension among participating GPUs. Then right before the attention computation, it employs all-to-all communication collective on the partitioned queries, keys and values such that each GPU receives the full sequence but only for a non-overlapping subset of the attention heads. This allows the participating GPUs to compute attention for different attention heads in parallel. Finally, DeepSpeed-Ulysses employs another all-to-all to gather the results along the attention heads while re-partitioning along the sequence dimension. 

In this work, we put forward the following contributions of DeepSpeed-Ulysses to advance state of the art in long sequence parallelism: 
\begin{itemize}
\item DeepSpeed-Ulysses trains Transformer models 4x larger sequence lengths than existing systems, while enabling training with sequences with over a million tokens.
\item Communication reduction of over 10x compared to existing systems, resulting in throughput improvements of up to 2.5x, and sustained throughput of over 175 TFlops/GPU (over 54\% of hardware peak).
\item Fully general and implementation agnostic attention: DeepSpeed sequence parallelism (Ulysses) supports dense as well as sparse attention, and it works with efficient attention implementations such as FlashAttention v2 ~\citep{dao2023flashattention2}. 
\item Support for massive model training: DeepSpeed sequence parallelism works together with ZeRO-3 to not only support large sequence lengths but also massive model sizes.
\item Easy-to-use and portable, requiring minimal code changes to the existing training frameworks.
\end{itemize}
In subsequent sections, we provide background and related work, a detailed discussion of DeepSpeed sequence parallelism core design, communication complexity analysis, experimental evaluation and comparison with existing work. 

\section{Background and Related Work}
In this section, we present a brief overview of Transformer architecture, mode of parallelism to accelerate Transformer training and a discussion on closely related work to our approach

\subsection{Background}
This section briefly introduces Transformer architecture and highlights different mode of parallelism of deep neural network in general and Transformer model in particular. This brief discussion is followed by specific focus on closely related work.
\subsubsection{Transformer Architecture}
\begin{figure*}[htb]
    \centering
    \includegraphics[width=.75\linewidth]{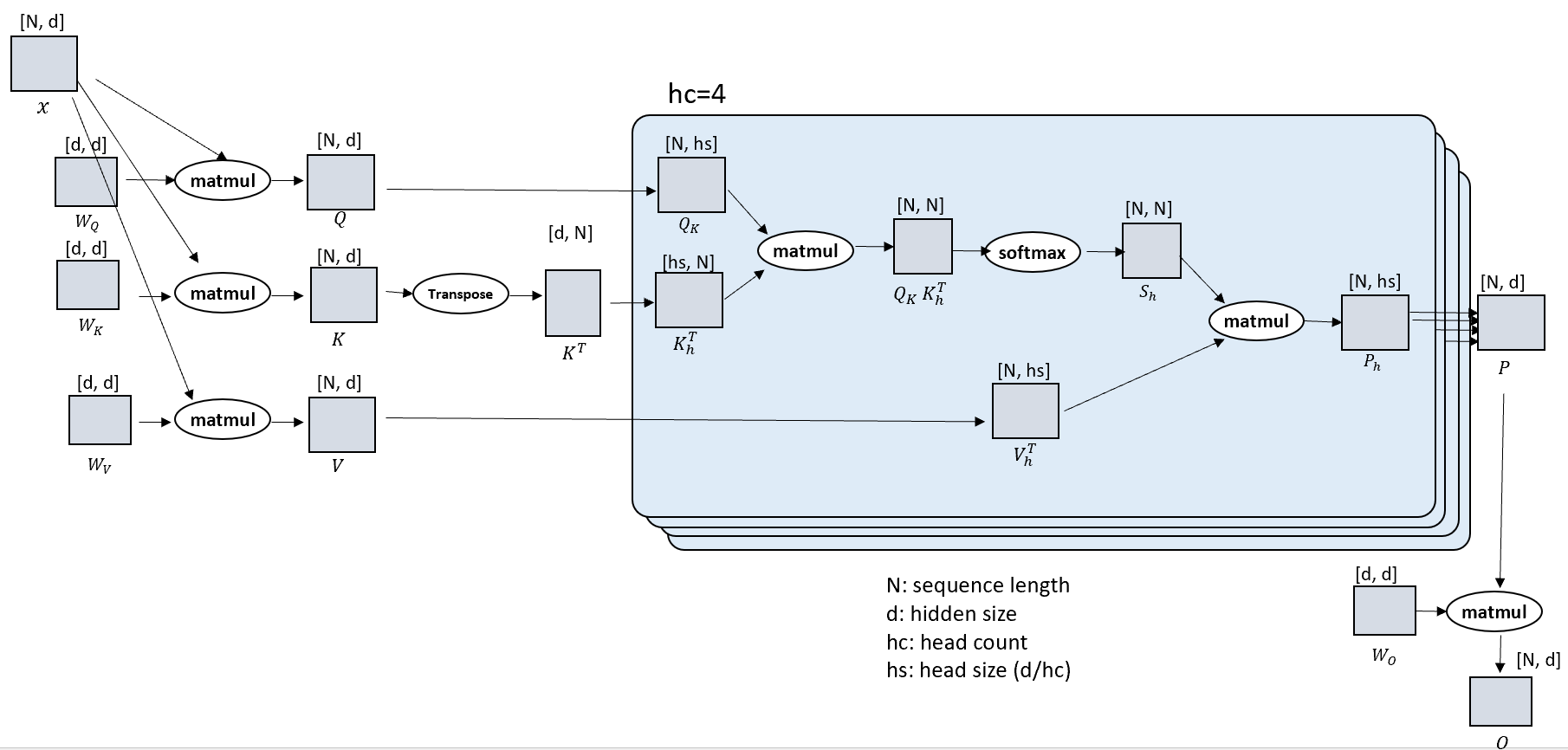}
    \caption{Multi-head attention Transformer}
    \label{fig:mha_basic}
\end{figure*}

Shown in Figure ~\ref{fig:mha_basic} is a sketch of building blocks of a typical multihead attention Transformer architecture ~\citep{vaswani2017attention}. It consists of input sequences which are projected into queries (\emph{Q}),keys (\emph{K}) and values (\emph{V}) embeddings. \emph{QKV} are typically a 3D tensor of size $N,b, d$ where $N$ is sequence length, $b$ is micro batch size and $d$ is hidden dimension. The $QKV$ tensors are fed to the attention block, a central component of Transformer model. Outputs of attentions are inputs to the multilayer perceptron (MLP) or position-wise feed-forward block of Transformer architecture. 

The attention block followed by MLP block are replicated multiple times to form an encoder, a decoder or an encoder-decoder Transformer network.

\subsubsection{Mode of Parallelism}
Data parallelism ~\citep{dean2012large} is de facto method of accelerating neural network training and has been applied widely with different neural network architectures and applications. Data parallelism in its simplest form partitions input data across sample or batch dimension while replicating model parameters across compute devices. Data parallelism is effective when the batch size is sufficiently large to hide communication cost in compute. However, it is limited when model is large and model parameter replication across devices is practically infeasible. ZeRO ~\citep{rajbhandari2020zero,rajbhandari2021zero} optimization addresses this problem by partitioning model parameters across available compute devices. Moreso, large batch is known to have impacts on model quality ~\citep{keskar2016large}. 

It is worth to note that our proposed approach is orthogonal to both data parallelism and ZeRO. Our proposed approach can be used with both methods. Also, by leveraging sequence parallelism to keep global batch size at reasonable size on large systems, we effectively ameliorate the impact of large batch size on model convergence. Sequence parallelism serves two purposes in this regard. First, sequence parallelism can accelerate time to solution for same (already explored) long sequence length; in other words, sequence parallelism reduces the iteration time proportional to additional compute resources. Second, sequence parallelism enables longer sequence training or continual pretraining where training context length gradually increase over time ~\citep{xiong2023effective}. Consider a real world scenario of large scale training on 1024 GPUs. The initial exploratory or pretraining set up of a (proxy) LLM has a sequence length of 8192 (8K), a micro batch size of 1 (thus, 8 million token global size) per GPU. A simple change to improve the quality of the pretrained model requires a change of sequence length from 8K to 32K, which would result in approximately 32 million global batch size. However, increasing the global batch size is not an option due to the negative impact on model quality. Therefore, sequence parallelism comes in handy as a system optimization technique with no requirement for laborious hyperparameter search. In this scenario, sequence parallelism allows for large batch sizes to be split across multiple GPUs without increasing the global batch size, regardless of the sequence length. 

Tensor ~\citep{megatronlm} and pipeline parallelism ~\citep{narayanan2019pipedream,GPipe,narayanan2021memory} are two other popular methods for large scale training. Collectively, tensor and pipeline parallelism are called model parallelism, and are targeted at compute operators in large models. In contrast to data parallelism, model parallelism are used when models are too large (as it is in many LLMs) and  can not be fully replicated across data parallel ranks.  Tensor parallelism splits compute operators (i.e., attention and MLPs) within a layer and pipeline parallelism splits model in a depth-wise (layer-wise) fashion. 3D parallelism ~\citep{deepspeed3dblog,smith2022using} combines data parallelism, tensor parallelism and pipeline parallelism to achieve higher throughput in comparison to the 3 constituents components at a cost of extensive code rewrite and productivity overhead ~\citep{wang2023zero}.

\subsection{Related Work}
For a broad overview and survey of distributed training methods for deep neural networks please see ~\citep{ben2019demystifying}. These methods are broadly categorized into data and model parallelism as described above. However, all of existing parallel methods are limited in dealing with intermediate activation memory overhead associated with extremely long sequence. 

While recent works in sequence parallelism address the memory overhead, they are lacking in communication efficiency, thus limited in scaling capability. Similar to our work, all existing works in sequence parallelism partition the input data along sequence dimension but differ in what input projections are partitioned and how partitions are aggregated and communicated for attention computation.

The authors in ~\citep{li2022sequence} (henceforward called \emph{ColAI-SP}) introduce ring self attention, a ring-like communication collective in which query projections are local whereas key and values projections are transmitted in a ring-style to compute global attention, resulting in communication complexity linear in message size, $M$. Megatron-LM sequence parallelism ~\citep{korthikanti2022reducing} approach is tightly integrated with Megatron tensor parallelism. Megatron LM partitions sequence along sequence dimensions and applies allgather and reduce scatter collective to aggregate \emph{QKV} projections for attention computation. Communication complexity analysis shows that unlike our approach, Megatron-LM sequence parallelism communication volume increase linearly with message size ($M$) regardless of number of compute devices. DeepSpeed-Ulysses on the other hand keeps communication volume consistent by increasing GPUs proportional to message size or sequence length see ~\ref{sec:design:comm} for more details. 

Table ~\ref{tab:sp-methods} summarizes how DeepSpeed-Ulysses differs from other existing methods. DeepSpeed-Ulysses has communication efficiency advantage over the other two methods. It also benefits from leveraging ZeRO ~\citep{rajbhandari2020zero,rajbhandari2021zero} optimization for model parameter partitioning across both sequence and data parallel groups. DeepSpeed-Ulysses supports different kinds of attention and it is easy to use. Megatron-LM sequence parallelism is tightly integrated with Megatron-LM tensor parallelism limiting both its memory efficiency and easy of use. \emph{ColAI-SP} requires a different (specific) kind of attention and is not easy to use. It is not clear how well \emph{ColAI-SP} ring self-attention generalizes to other attention types and mechanisms.

\begin{table}[t]\footnotesize\fontsize{6.75}{\baselineskip}\selectfont
    \centering
    \setlength{\tabcolsep}{1.5pt}
    \begin{tabular}{cccccc}
         \toprule
         \multirow{2}{*}{Method} & Comm & Activation & Parameter & Attention & Ease \\
         & complexity & memory efficiency & memory efficiency & agnostic & of use \\
         \midrule
         ColAI-SP ~\citep{li2022sequence} & $O(M)$ & \greencheck{} & \redx{} & \redx{} & \redx{} \\
         Megatron-SP ~\citep{korthikanti2022reducing} & $O(M)$  & \greencheck{}  & \redx{} & \greencheck{} & \redx{}  \\
         \textbf{DS-Ulysses} & $O(M/P)$ & \greencheck{} & \greencheck{} & \greencheck{} & \greencheck{}\\
         \bottomrule
    \end{tabular}
    \caption{Comparison of our work (DS-Ulysses) to other sequence parallelism methods.}
    \label{tab:sp-methods}\vspace{-3em}
\end{table}

There are related works in sparse Transformer particularly focusing on full-attention approximation such as sparse attention ~\citep{sparse-attention,DBLP:journals/corr/abs-2009-14794,zaheer2021big,beltagy2020longformer}. There are also recent works on single GPU memory and compute efficient attention. A popular example in this category is Flash attention ~\citep{dao2022flashattention,dao2023flashattention2}, which leverages known techniques such as tiling and recomputation for compute and memory efficiency. These works are orthogonal to our work and were leveraged accordingly. 
\section{DeepSpeed-Ulysses Core Design}
\subsection{System Design}
\label{sec:design}
\begin{figure*}[htb]
    \centering
    \includegraphics[width=.75\linewidth]{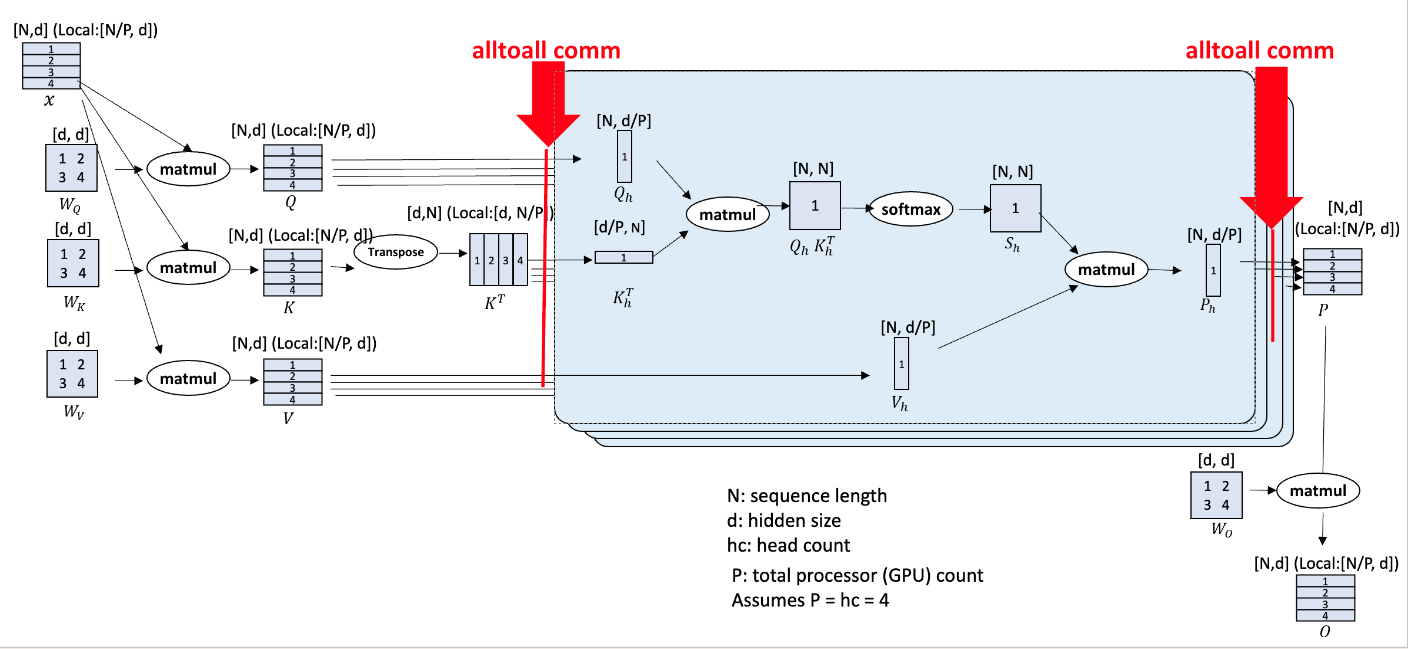}
    \caption{DeepSpeed sequence parallelism (DeepSpeed-Ulysses) design}
    \label{fig:ds_seq_design}
\end{figure*}
Figure ~\ref{fig:ds_seq_design} shows the core design of DeepSpeed-Ulysses. As with the known transformer architecture, the design consists of input sequences \emph{N} partitioned across \emph{P} available devices. Each local \emph{N/P} partition is projected into queries (\emph{Q}), keys (\emph{K}) and values (\emph{V})  embeddings. Next, (\emph{QKV}) embeddings are gathered into global \emph{QKV} through highly optimized all-to-all collectives between participating compute   devices. Sequel to all-to-all  collective is the attention computation per head in the form: 
\begin{equation}\label{eq:attn}
Output context=Softmax ((QK^T)/\sqrt(d))V
\end{equation}

After the attention computation, another all-to-all collective transforms output context tensor of attention computation to sequence (\emph{N/P}) parallel for subsequent operators (MLP MatMul, layer norm etc) in the remaining modules of transformer layer block. 

\subsection{Communication Analysis}
\label{sec:design:comm}
What distinguishes DeepSpeed-Ulysses from the other existing long-sequence approaches is our much smaller aggregate communication volume and overall better scalability with increasing degree of sequence parallelism compared to existing solutions, as demonstrated by the communication volume analysis below: 

On modern clusters with intra-node NVSwitch interconnect and inter-node fat tree IB topology, the communication volume transmitted per link for an all-to-all for aggregate message of size \emph{M} over \emph{P} GPUs is \emph{M/P}. For a transformer model with hidden size h, sequence length of N, and parallelism degree of P, DS-Sequence performs all-to-all for the \emph{QKV} projections with an aggregate message size of \emph{3Nh} before the attention computation, and another all-to-all for output context projection with a size \emph{Nh} for each transformer layer.  Therefore, DeepSpeed sequence parallelism incurs an aggregate communication volume per link of \emph{4Nh/P} (or with the complexity of \emph{O(N/P)}. Note that this communication volume is constant when both \emph{N} and \emph{P} are increased proportionally. 

 In contrast, the existing approaches like Megatron-LM incur communication volume that increases linearly with N regardless of P, resulting in the communication complexity of \emph{O(N)}.  For instance, Megatron-LM performs two all-gather with the message volume of \emph{Nh} and two reduce-scatter with the volume of \emph{Nh} for each transformer layer. However, the cost of each all-gather and reduce-scatter of size \emph{M} remains \emph{M} when \emph{P >> 1}, instead of \emph{M/P}. Therefore, Megatron-LM sequence parallelism incurs a communication volume per link of \emph{4Nh} which is \emph{P} times larger than that for DeepSpeed sequence parallelism. This allows DeepSpeed sequence parallelism  to enable training with extremely long sequences while achieving significantly higher training efficiency compared to the existing approaches. Our evaluation results match this analysis. 

\subsection{Memory Efficiency}
\label{sec:design:mem}
While DeepSpeed sequence parallelism reduces the activation memory when training with longer sequences, it does not impact the memory consumed by the model states. Therefore, to support large sequence length training with a large language model, DeepSpeed sequence parallelism is integrated with ZeRO-3.
ZeRO Redundancy Optimizer Stage 3 (ZeRO-3) \citep{rajbhandari2020zero,rajbhandari2021zero} is a memory optimization technique for training large models. Unlike the classic data parallel training of neural networks where model states are replicated across data parallel ranks, ZeRO-3 optimizes memory usage by partitioning model states across data parallel ranks. However, with sequence parallelism, training data can be considered in both batch (sample) and sequence dimensions and the associated parallel groups combined to form a larger group for ZeRO parallelism.
Therefore, we extend ZeRO-3 partitioning to combination of data parallel and sequence parallel ranks. In other words, in DeepSpeed sequence parallelism, ZeRO partitions model states across both sequence and data parallel group and collects per rank partitions (allgather) when they are needed. Similarly, gradients are reduced across both data and sequence parallel ranks for parameter update. ZeRO support allows for huge memory savings in both sequence and data dimensions and enables scaling not just to large sequence lengths but also to large models.     

\subsection{General and Attention Agnostic Solution}
\label{sec:design:use}
DeepSpeed implementation of distributed attention module is general enough to support any attention: e.g., self-attention, cross-attention, causal attention in both their dense and sparse counterparts, and their various optimized kernels that support long-sequence at local attention level such as different versions of FlashAttention. 
The general property of DeepSpeed-Ulysses stems from the modular nature of its core design: an attention-centric sequence parallelism design. Prior to attention computation is sequence parallelism of N/P partition, attention computation is head parallelism with full attention per head but just with fewer heads, thus attention computation can be replaced with any type of attention mechanisms, e.g., dense attention and various forms of sparse attention.

\section{Evaluation}
We evaluate DeepSpeed-Ulysses (DeepSpeed Sequence) on GPT \citep{gpt-2}, a foundation model for many NLP tasks on up to 256 A100 GPUs. Our evaluations are five-fold:  i) sequence length scalability,  ii) throughput for dense attention and comparison with existing system, and iii) throughput with sparse attention and comparison with existing system, iv) parallel scaling study and v) convergence study of Deep sequence parallelism. We discuss and present evaluations from each of these categories next.

\subsection{Sequence Length Scalability}
\begin{figure}[h]
    \centering
    \includegraphics[width=.8\linewidth]{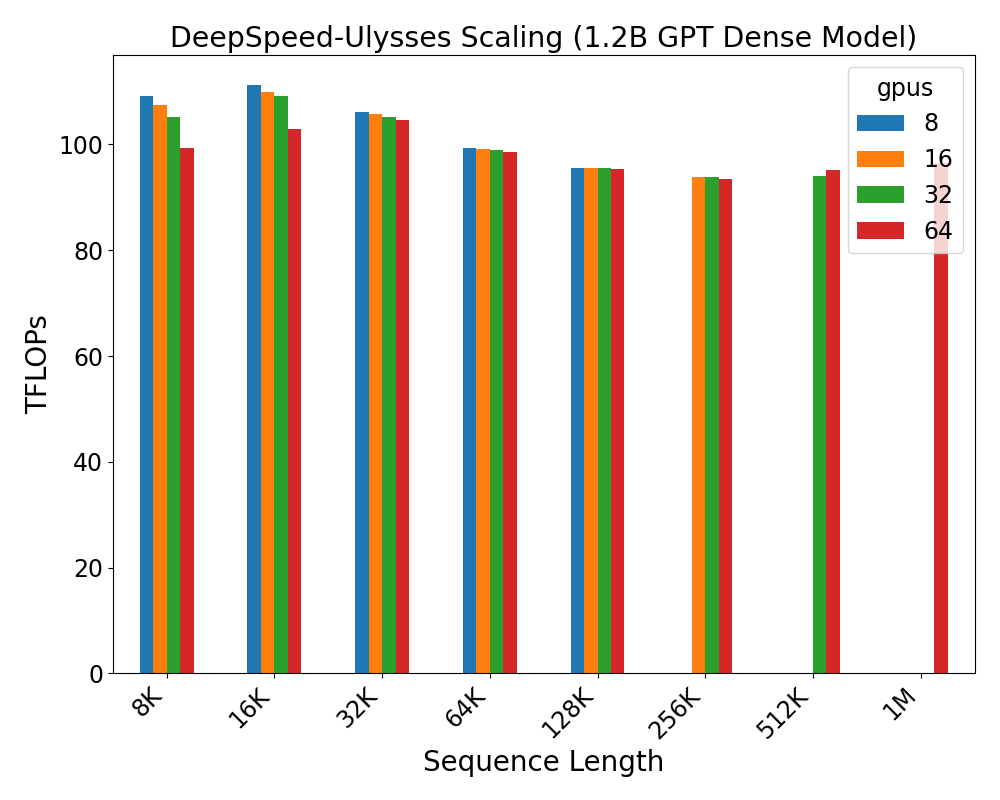}
    \caption{DeepSpeed sequence parallelism strong scalability evaluation at different sequence length and GPU counts}
    \label{fig:seqlen-scale}
\end{figure}

The first set of experiments is strong scaling of sequence length up to 1 million tokens on 1.2 billion parameter GPT model. Results of this evaluation are shown in Figure~\ref{fig:seqlen-scale}. DeepSpeed sequence parallelism allows increasing sequence length linearly with the number of GPUs and sequence length scales linearly relative to and maintains similar computation throughput across different sequence length at appropriate GPU count.

\subsection{Dense Attention Evaluation}
\begin{figure}[h]
    \centering
    \includegraphics[width=.8\linewidth]{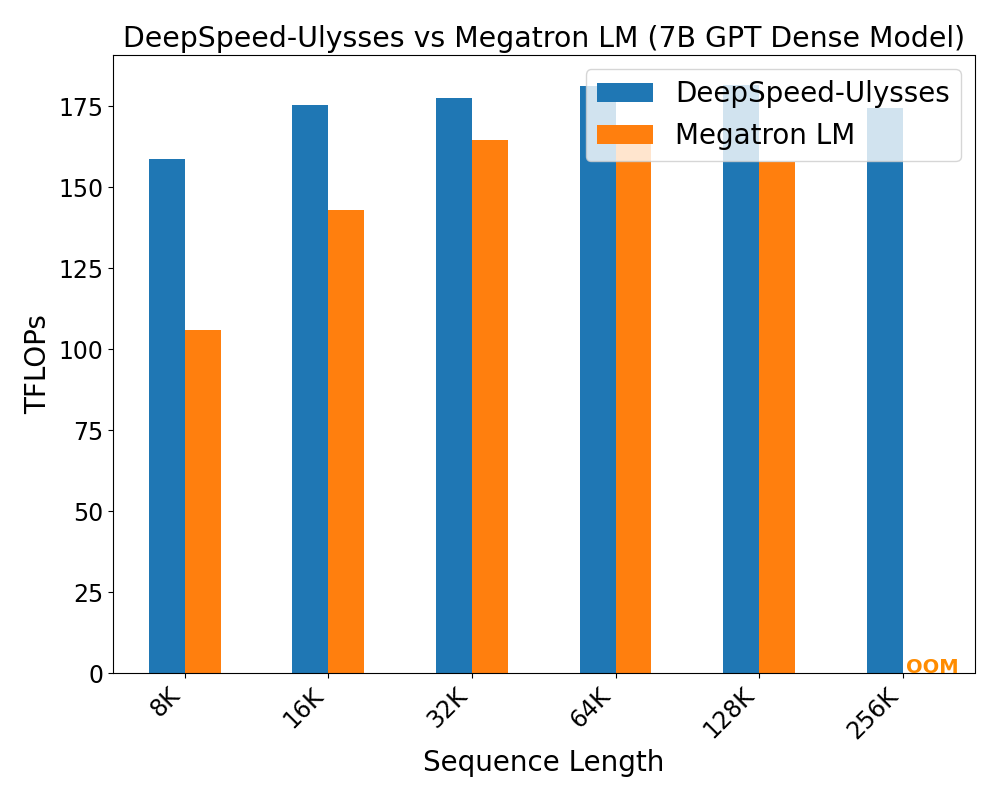}
    \caption{Evaluation of DeepSpeed-Ulysses and Megatron LM on 7B parameter model with dense attention (32 GPUs)}
    \label{fig:dense_7b}
\end{figure}

\begin{figure}[h]
    \centering
    \includegraphics[width=.8\linewidth]{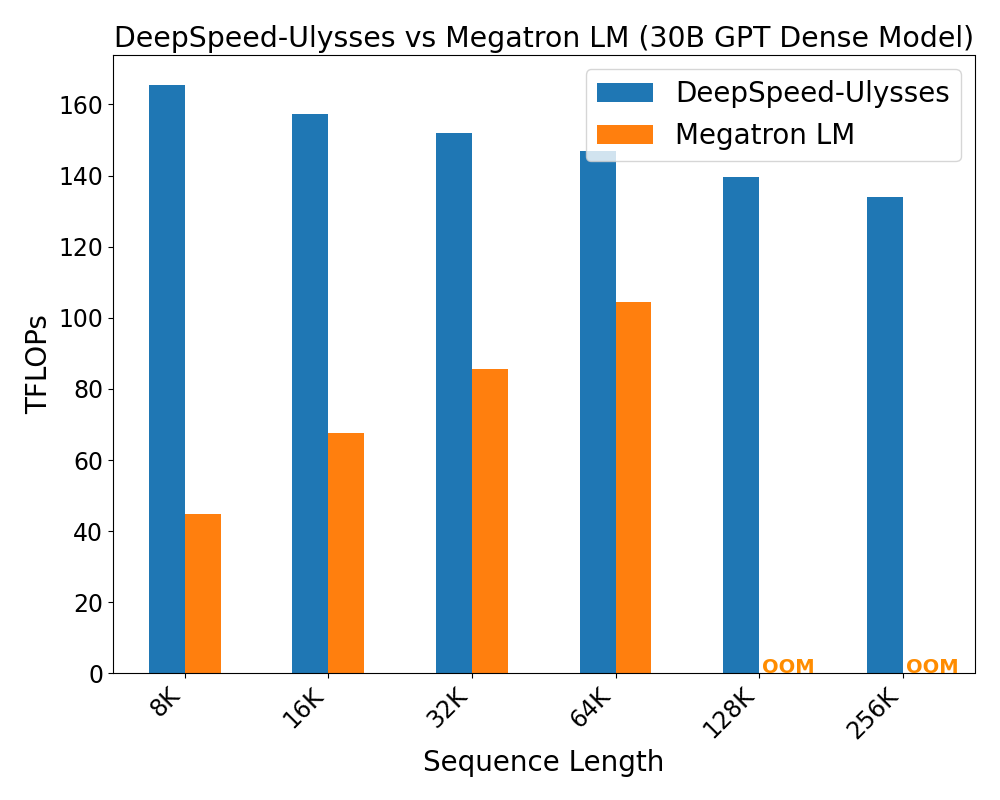}
    \caption{Evaluation of DeepSpeed-Ulysses and Megatron LM on 30B parameter model with dense attention (64 GPUs)}
    \label{fig:dense_30b}
\end{figure}

Next, we evaluate DeepSpeed sequence parallelism on 7 billion (7B) and 30 billion (30B) parameter GPT dense attention models and compare against Megatron-LM's sequence parallelism on 32 and 64 A100 GPUs respectively. The results of these evaluations are shown in Figures~\ref{fig:dense_7b} and ~\ref{fig:dense_30b}.  

We compare DeepSpeed sequence parallelism with Megatron-LM for 7B and 30B models running various sequence lengths. For our evaluation we chose the sequence parallelism degree and micro-batch size that produced the best performance (measured as throughput or TFLOPs) for both DeepSpeed sequence parallelism and Megatron-LM, this we call optimal (batch size-sequence length) configurations. For DeepSpeed sequence parallelism, we always use a ZeRO parallelism degrees of 32 and 64 for 7B and 30B models respectively.

Figures ~\ref{fig:dense_7b} and ~\ref{fig:dense_30b} show that DeepSpeed sequence parallelism consistently outperforms Megatron-LM for the sequence length that can be run with both. In addition, DeepSpeed sequence parallelism can run longer sequence than Megatron-LM. DeepSpeed sequence parallelism performance advantages are two folds: (1) DeepSpeed sequence parallelism in combination with ZeRO-3 fits more samples than Megatron-LM because of the memory optimization leading to higher throughput (2) DeepSpeed sequence parallelism benefits from efficient all-to-all communication relative to all-gather communication as applied in Megatron-LM sequence parallelism. 

\subsection{Sparse Attention Evaluation}
\begin{figure}[h]
    \centering
    \includegraphics[width=.8\linewidth]{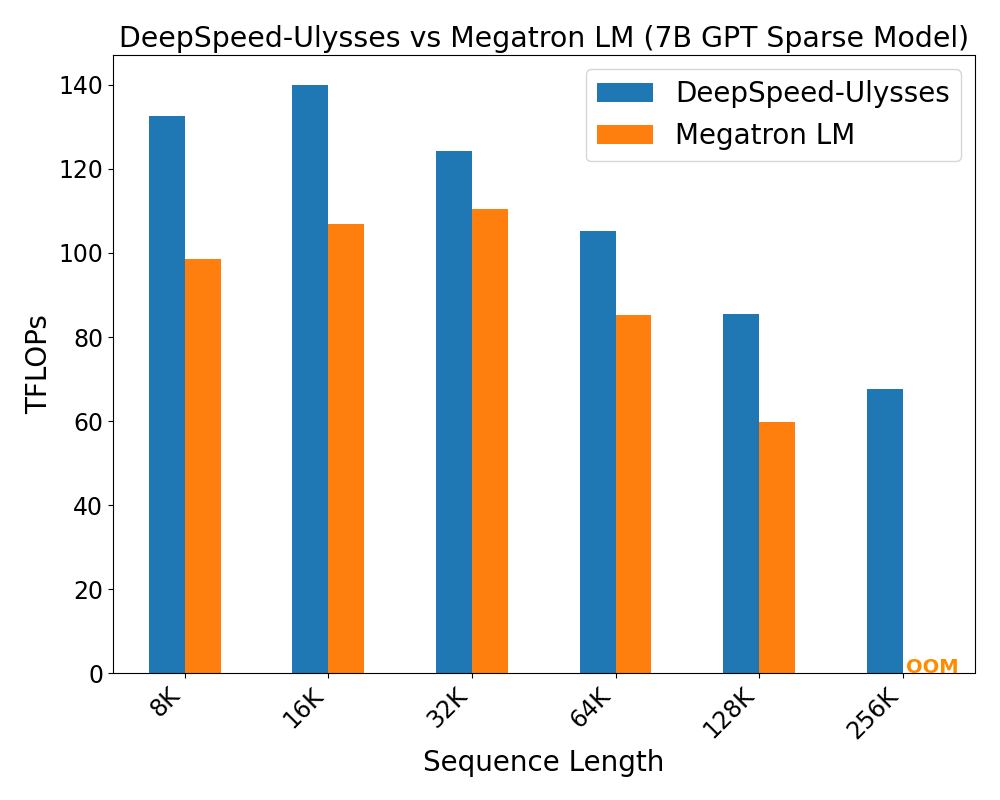}
    \caption{Evaluation of DeepSpeed-Ulysses and Megatron LM on 7B parameter model with blocked sparse attention (32 GPUs)}
    \label{fig:sparse_7b}
\end{figure}

\begin{figure}[h]
    \centering
    \includegraphics[width=.8\linewidth]{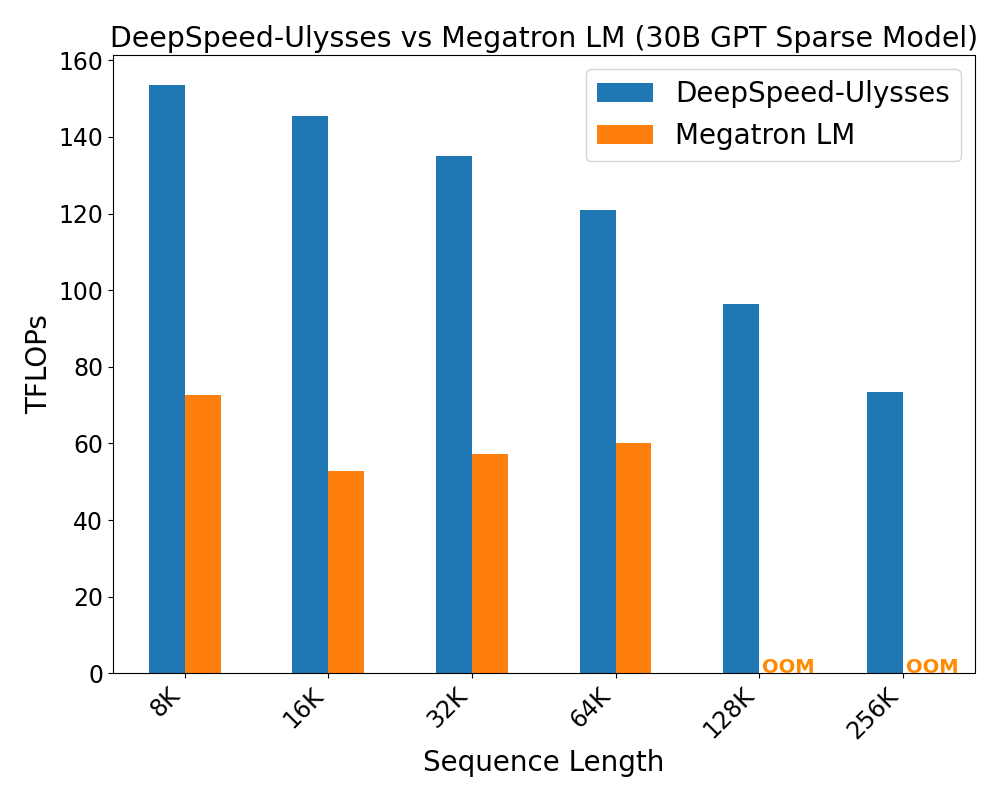}
    \caption{Evaluation of DeepSpeed-Ulysses and Megatron LM on 30B parameter model with blocked sparse attention (64 GPUs)}
    \label{fig:sparse_30b}
\end{figure}

Similarly, we evaluate DeepSpeed sequence parallelism on 7 billion and 30 billion parameter sparse attention models and benchmark against Megatron-LM sequence parallelism. Results of our evaluation are shown in Figures ~\ref{fig:sparse_7b} and ~\ref{fig:sparse_30b}. We observe similar trends with sparse attention as dense attention experiments. We observe more than 2x throughput performance of DeepSpeed sequence parallelism compared to Megatron-LM. For memory saving, DeepSpeed sequence parallelism leveraging ZeRO-3 scales to 4x longer sequence lengths than Megatron-LM. 

DeepSpeed sequence parallelism outperforms Megatron-LM for sequence length that can be run with both. In fact, the current DeepSpeed throughput is bottlenecked by the local sparse attention implementation, and as a result DeepSpeed throughput decreases as the sequence length increases. We expect this gap in performance between DeepSpeed and Megatron-LM to increase further for larger sequence lengths as we improve the performance of the local sparse attention implementation in future.

\subsection{Parallel Scaling Study}
\begin{table}[h]\footnotesize\fontsize{8.85}{\baselineskip}\selectfont
    \centering
    \caption{Parallel scaling study with fixed sequence length}
    \label{tab:strong_scaling}
    \begin{tabular}{cccc}
        \toprule
        Seqlen & GPUs & Time (ms) & TFLOPs\\
        \midrule
        131072 & 64 & 32432.1333 &165.526667\\
        131072 & 128  &17052.5143  & 157.41\\
       131072  & 256  & 9886.7 & 136.09\\
       \bottomrule
    \end{tabular}
\end{table}

\begin{table}[h]\footnotesize\fontsize{8.85}{\baselineskip}\selectfont
    \centering
    \caption{Parallel scaling study with varying sequence length}
    \label{tab:weak_scaling}
    \begin{tabular}{cccc}
        \toprule
        Seqlen & GPUs & Time (ms) & TFLOPs\\
        \midrule
        65536 & 64 & 9676.76 &161.3626667\\
        131072 & 128  &17052.5143  & 157.41\\
       262144  & 256  & 33486.5 & 147.4\\
       \bottomrule
    \end{tabular}
\end{table}

Furthermore, we conduct parallel scaling studies of DeepSpeed-Ulysses along two axes. First, we fix sequence length at 131,072 tokens and increase GPU count from 64 to 256. Second, we increase the GPU count proportionally to the increase in sequence length. The results of these experiments are shown in Tables ~\ref{tab:strong_scaling} and ~\ref{tab:weak_scaling} respectively. For both evaluations, we used GPT-7B dense model at global batch size of 8. The tables show iteration time in microseconds as well as the achieved throughput measured in per GPU TFLOPs. Table~\ref{tab:strong_scaling} can be interpreted as strong scaling and shows that execution time decreases almost linearly as we increase the GPU count. Table ~\ref{tab:weak_scaling} on the other hand, is a form of weak scaling (not in the traditional sense) with caveat that attention computation, a function of sequence length, is quadratic in complexity. In other words, as we increase sequence length, the work increases quadratically. 

Communication overhead can be attributed to slight decrease in throughput as we increase communication workload (that is, sequence length or GPU count). This overhead notwithstanding, we observe good scaling at high percentages of theoretical peak GPU performance across the two studies. These good scaling results indicate good parallel efficiency of DeepSpeed-Ulysses.

\subsection{Convergence Study}
\begin{figure}[h]
    \centering
    \includegraphics[width=.8\linewidth]{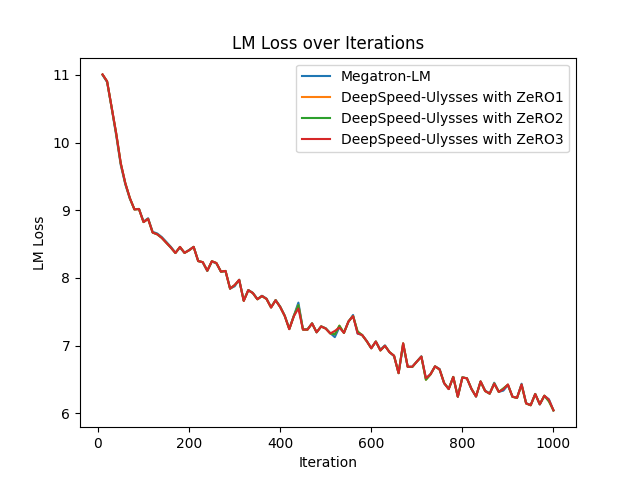}
    \caption{Convergence evaluation of DeepSpeed-Ulysses with different ZeRO memory optimization stages}
    \label{fig:convg}
\end{figure}

Lastly, Figure ~\ref{fig:convg} shows convergence of a 1.3 billion GPT model at 32K sequence length on 8 A100 GPUs with sequence parallelism degree set at 4 for both DeepSpeed-Ulysses and Megatron-LM sequence parallelism.  For DeepSpeed sequence parallelism, we evaluate convergence with different ZeRO stages.  DeepSpeed sequence parallelism is a purely system optimization technique that enables training of long sequence Transformer model, thus there is no (negative) on quality of trained models, this assertion is validated through experiments and is shown in Figure ~\ref{fig:convg}.

\section{Conclusion}
In conclusion, we present a memory and communication efficient DeepSpeed Sequence as enabling technology for long sequence large Transformer training. DeepSpeed Sequence enables sequence parallelism across GPUs (by extension other AI accelerators), parallelizing sequence across all components of the Transformer model, including streamline support for SOTA Flash (dense and sparse) attention. Training with DeepSpeed Sequence allows both model size and sequence length to scale near indefinitely unbounded by single GPU memory limitation and at a high fraction of peak compute performance. 

\bibliographystyle{unsrtnat}
\bibliography{main} 

\end{document}